\documentclass{article} 
\usepackage{iclr2026_conference,times}

\usepackage{amsmath,amsfonts,bm}

\def\eqref#1{equation~\ref{#1}}

\def\1{\bm{1}}

\def\vx{{\bm{x}}}

\DeclareMathAlphabet{\mathsfit}{\encodingdefault}{\sfdefault}{m}{sl}
\SetMathAlphabet{\mathsfit}{bold}{\encodingdefault}{\sfdefault}{bx}{n}

\def\sR{{\mathbb{R}}}

\usepackage{hyperref}
\usepackage{url}

\usepackage{graphicx}
\usepackage{subcaption}
\usepackage{float}   
\usepackage{booktabs} 
\usepackage{multirow}
\usepackage{amssymb}
\usepackage{ulem}  
\usepackage{fontawesome5} 
\usepackage{microtype}  
\usepackage[scaled=1.0]{inconsolata}  

\usepackage{bm}

\def\eqref#1{equation~\ref{#1}}

\def\1{\mathbf{1}}

\def\vx{{\mathbf{x}}}

\DeclareMathAlphabet{\mathsfit}{\encodingdefault}{\sfdefault}{m}{sl}
\SetMathAlphabet{\mathsfit}{bold}{\encodingdefault}{\sfdefault}{bx}{n}

\def\sR{{\mathbb{R}}}

\usepackage{cancel}

\newcommand{\norm}[1]{\left\lVert#1\right\rVert}

\newcommand{\lockicon}{\scalebox{0.85}{\faLock}}
\newcommand{\unlockicon}{\textcolor[HTML]{6c3baa}{\scalebox{0.85}{\faUnlock}}}

\title{On the Value of Tokeniser Pretraining in Physics Foundation Models}

\author{
\textbf{Hadi Sotoudeh}\textsuperscript{\textnormal{1}}\thanks{Contact: \texttt{mhs60@cam.ac.uk}}
\and
\textbf{Payel Mukhopadhyay}\textsuperscript{1}
\and
\textbf{Ruben Ohana}\textsuperscript{2}
\and
\textbf{Michael McCabe}\textsuperscript{2,3}
\and
\textbf{Neil D. Lawrence}\textsuperscript{1}
\qquad\quad
\textbf{Shirley Ho}\textsuperscript{2,3,4}
\qquad\quad
\textbf{Miles Cranmer}\textsuperscript{1}
\\[6pt]
The Polymathic\,AI Collaboration\\
\textsuperscript{1}\,University of Cambridge \quad
\textsuperscript{2}\,Flatiron Institute \quad
\textsuperscript{3}\,New York University \quad
\textsuperscript{4}\,Princeton University
}

\iclrfinalcopy 
\begin{document}

\maketitle

\begin{abstract}
We investigate the impact of tokeniser pretraining on the accuracy and efficiency of physics emulation. Modern high-resolution simulations produce vast volumes of data spanning diverse physical regimes and scales. Training foundation models \linebreak[4] to learn the dynamics underlying such data enables the modelling of complex multiphysics phenomena, especially in data-limited settings. The emerging class of physics foundation models typically aims to learn two tasks jointly: \linebreak[4] (i) extracting compact representations of high-resolution spatiotemporal data, and (ii) capturing governing physical dynamics. However, learning both tasks from scratch simultaneously can impede the effectiveness of either process.
We show that pretraining the tokeniser with an autoencoding objective prior to training the dynamics model enhances computational efficiency for physics emulation. Notably, the magnitude of this benefit depends on domain alignment: pretraining on the same physical system as the emulation task yields the largest improvements, while pretraining on other systems provides moderate gains. In-domain pretraining reduces VRMSE by 64\% after 10,500 training steps compared to training from scratch.
\linebreak[4] To our knowledge, this is the first systematic investigation of tokeniser pretraining for physics foundation models.
We further introduce flexible spatiotemporal compression operations that extend causal convolutions to support runtime-adjustable compression ratios, enabling efficient adaptation to diverse downstream tasks. \linebreak[4] Our findings provide practical guidance for training efficient physics emulators and highlight the importance of strategic pretraining data selection.
\end{abstract}

\section{Introduction}

Modelling physical systems stands to benefit significantly from large-scale pretraining.
Foundation~models trained across diverse physical systems can capture shared patterns that span different physical regimes and scales; patterns that are often beyond the scope of any single simulation. However, the high spatial and temporal resolution of scientific data makes it computationally prohibitive to train large-scale transformer-based foundation models directly in pixel space. This underscores the need for efficient approaches to processing such data.

In state-of-the-art generative systems operating on high-resolution data, the modelling process is often decomposed into two stages: (i) \textbf{Tokenisation:} extracting compact, informative representations of temporally and spatially high-frequency features using convolutional layers,\footnote{Here, ``tokenisation'' refers to learning representations (i.e., tokens) from data that will be used to train another network; it does not imply that the tokens are quantised or discrete.} and (ii) \textbf{Prediction:} feeding these representations into a transformer-based network to model global dependencies and make predictions \citep{nvidia2025_cosmos, yu2024_magvit2, tschannen2024_givt, yu2023_magvit, chang2022_maskgit}.

This separation is motivated by the hypothesis that low-level (i.e., high-frequency) structure is best captured by locally connected architectures -- namely, convolutions. At higher semantic levels (i.e., lower frequencies), global interactions become more important, making transformer architectures more suitable \citep{esser2021_vqgan}. This hypothesis is supported by observations that transformers tend to learn convolution-like structures at small scales \citep{dosovitskiy2021_vit}, motivating the explicit use of convolutional layers for encoding high-frequency information. Convolutions are also significantly more efficient to train than transformers due to their local connectivity, parameter~sharing, and linear scaling with input size. A growing body of work supports training the tokeniser and transformer in two separate stages rather than jointly from scratch, and this staged approach has become standard in high-resolution image and video generation \citep{nvidia2025_cosmos, yu2024_magvit2, rombach2022_diffusion}.

Recent physics foundation models \citep{mccabe2025_walrus, mccabe2024_mpp, herde2024_poseidon, hao2024_dpot, pathak2022_fourcastnet} have achieved impressive results by training tokenisers and dynamics models jointly from scratch. However, this contrasts sharply with common practice in computer~vision, where pretrained tokenisers have become standard. This raises a natural question: can physics~foundation~models benefit from tokeniser pretraining in the same way?

Pretraining the tokeniser offers several potential advantages. First, it decouples representation learning from dynamics modelling, allowing each task to be learned more effectively. Second, a pretrained tokeniser can be reused across multiple downstream applications that share the same latent space, amortising the pretraining cost across different physics problems.

\textbf{Our contribution.} To our knowledge, this is the first systematic study of tokeniser pretraining for physics foundation models. We demonstrate that pretraining the tokeniser improves training efficiency for autoregressive prediction of physical dynamics, with benefits dependent on domain alignment between pretraining data and autoregressive training data. Specifically, in-domain pretraining (same physical system) yields larger improvements than out-of-domain pretraining (different systems). Our focus is on understanding the relative benefits of different pretraining strategies. We therefore design our experiments to facilitate systematic comparisons across pretraining conditions.

Additionally, we introduce flexible spatiotemporal compression operations that extend causal convolutions to support runtime-adjustable compression ratios. This enables the computational cost of downstream tasks to be adjusted dynamically by varying the temporal and spatial coarseness of tokens. Importantly, different physical systems exhibit varying degrees of compressibility -- for example, galaxy images compress more readily than turbulent fluid simulations -- and flexible compression accommodates this diversity, enabling broader applicability and better adaptation to diverse downstream tasks without retraining.

The rest of this paper is organised as follows: Section~\ref{sec:methods} describes our methodology, including the datasets, evaluation metrics, model design, and training procedures. Section~\ref{sec:results} presents our results on the impact of tokeniser pretraining in both in-domain and out-of-domain settings. Section~\ref{sec:discussion} discusses the implications of our findings, practical considerations for training physics foundation models, and directions for future work. Section~\ref{sec:conclusion} provides concluding remarks. Appendix~\ref{appendix:data} provides detailed dataset descriptions, Appendix~\ref{appendix:architecture} describes the tokeniser architecture and flexible compression operations, and Appendix~\ref{appendix:training} contains complete training hyperparameters.

\section{Methods} \label{sec:methods}

This section is structured as follows:\, First, we describe the datasets used in our experiments (Subsection~\ref{subsec:data}). Next, we outline the experimental design and evaluation metrics (Subsection~\ref{subsec:experiment_setup}). We then present the training objectives for both the dynamics model and tokeniser (Subsection~\ref{subsec:objective}), followed by introducing the model architectures (Subsection~\ref{subsec:architecture}). Finally, we detail the parameter counts and training procedures (Subsection~\ref{subsec:training}).

\subsection{Data} \label{subsec:data}

We use \textit{The Well} \citep{ohana2025_thewell}, a large-scale collection of diverse physics simulations designed for machine learning applications. From this collection, we use the following 2D datasets: Euler multiquadrants, Rayleigh-Bénard convection, shear flow, and active matter (see Appendix~\ref{appendix:data} for details). We divide simulated trajectories into 10-frame sequences. Each sequence contains multiple physical fields (e.g., velocity, pressure, density), which vary by dataset. Dataset properties are summarised in Table~\ref{tbl:data}.

\begin{table}[h]
\caption[Dataset summary]{Dataset sizes, dimensions, and available fields.}
\centering
\label{tbl:data}
\small
\begin{tabular}{l l c c c c}
\toprule
& & \multicolumn{4}{c}{Dataset} \\
\cmidrule(lr){3-6}
& Training size & 736k & 267.4k & 171.1k & 12.6k \\
& Dimensions & (512,512) & (512,128) & (256,512) & (256,256) \\
\midrule
\textbf{Field Type} & \textbf{Field Name} & \textbf{Euler} & \textbf{Rayleigh-Bénard} & \textbf{Shear Flow} & \textbf{Active Matter} \\
\midrule
\multirow{6}{*}{Scalar}
& Energy        & \checkmark &           &           &           \\
& Density       & \checkmark &           &           &           \\
& Pressure      & \checkmark & \checkmark & \checkmark &          \\
& Buoyancy      &           & \checkmark &           &           \\
& Tracer        &           &           & \checkmark &           \\
& Concentration &           &           &           & \checkmark \\
\midrule
\multirow{2}{*}{Vector}
& Momentum & \checkmark &           &           &           \\
& Velocity &           & \checkmark & \checkmark & \checkmark \\
\midrule
\multirow{2}{*}{Tensor}
& Orientation &           &           &           & \checkmark \\
& Strain-Rate &           &           &           & \checkmark \\
\bottomrule
\end{tabular}
\end{table}

\subsection{Experiments and Evaluation Criteria} \label{subsec:experiment_setup}

\begin{figure}[H]
    \centering    
    \includegraphics[width=0.95\textwidth]{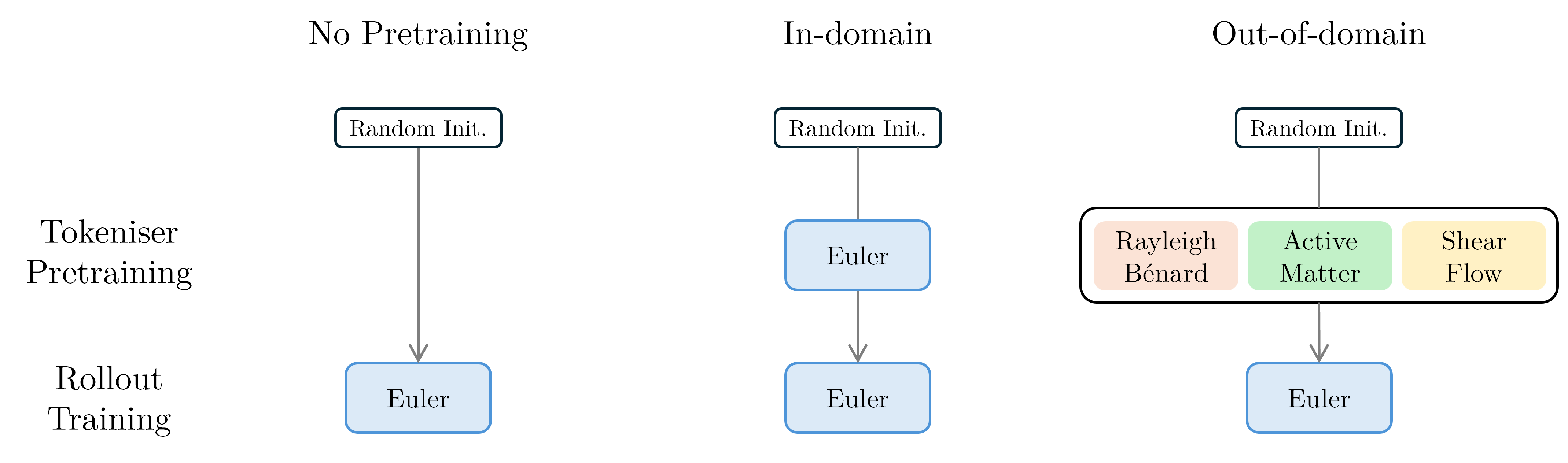}
    \caption[Experiments]{Overview of the experimental setup.}
    \label{fig:setup}
\end{figure}

The primary objective of this study is to investigate whether pretraining the tokeniser provides measurable benefits for foundation model training. To this end, we conduct controlled experiments comparing autoregressive rollout models trained with pretrained tokenisers against those with randomly initialised tokenisers (Figure~\ref{fig:setup}).

For the pretrained models, we consider two settings: in-domain, where both the tokeniser and rollout model are trained on the same dataset (Euler), and out-of-domain, where the tokeniser is pretrained on a mixture of other datasets (Rayleigh-Bénard, active matter, and shear flow) before the rollout model is trained on Euler. The latter tests whether the benefits of pretraining generalise across physical systems. Aside from the tokeniser, the rest of the rollout model is randomly initialised in all settings.

We evaluate the impact of tokeniser pretraining on \textbf{training cost} and \textbf{model performance}, asking: What is the gain, in terms of compute or quality, when using a pretrained tokeniser? Specifically, does pretraining reduce the computation required to reach a target performance level, or improve model quality under a fixed computational budget? We focus our analysis on early stages of downstream training, where practitioners with limited computational resources would most benefit from training efficiency improvements. Training cost is quantified in wall-clock time. We evaluate model performance using two complementary metrics:

\textbf{Spatial error (VRMSE).} We assess reconstruction quality using variance-normalised root mean squared error:
\begin{align}
\mathrm{VRMSE} = \frac{\sqrt{\mathbb{E}_{p}\!\bigl[( x_{p}-\hat{x}_{p} )^{2} \bigr] }}{\sigma_{\vx}},
\end{align}
where $\vx$ and $\hat{\vx}$ denote the target and predicted fields, $p$ indexes pixels, and $\sigma_\vx$ is the standard deviation of the target field computed over all pixels. This measures reconstruction error relative to the variability in the~target. We compute VRMSE per timeframe and per field, then average across both to obtain aggregate metrics.

\textbf{Spectral error (normalised error power spectrum).} Spatial metrics do not reveal how well a model captures structure at different scales. We therefore examine spectral errors by analysing the power spectrum of the residual field. For fields $\vx$ and $\hat{\vx}$, we compute the power spectrum of the residual~$\vx - \hat{\vx}$ by binning Fourier modes $\mathbf{k}'$ according to their wavenumber magnitude $|\mathbf{k}'|$ and averaging the squared Fourier coefficients within each bin $B_k$:
\begin{align}
P_{\vx - \hat{\vx}}(k) = \frac{1}{|B_k|} \sum_{\mathbf{k}' \in B_k} \left| \mathcal{F} \{ \vx - \hat{\vx} \} ( \mathbf{k}' ) \right|^2 .
\end{align}
This power spectrum quantifies the contribution of each wavenumber band to the total pixel-space MSE via Plancherel's theorem: the total MSE equals $\frac{1}{N^2}\sum_k |B_k| \, P_{\vx - \hat{\vx}}(k)$, where $N^2$ is the number of spatial pixels. To obtain a scale-invariant metric, we normalise by the true field's power spectrum at each scale:
\begin{align}
\text{NEPS}(k) = \frac{P_{\vx - \hat{\vx}}(k)}{P_{\vx}(k)},
\end{align}
where $P_{\vx}(k)$ is the power spectrum of the true field, computed analogously to $P_{\vx - \hat{\vx}}(k)$. This normalised error power spectrum (NEPS) measures the ratio of error power to signal power at each scale. Since each frequency band uses its own normalisation, NEPS values are not additive across scales but provide a relative quality metric independent of the natural energy distribution.
A value of NEPS$(k) = 0.1$ indicates the error power is 10\% of the signal power at scale $k$, while NEPS$(k) = 1.0$ means the error power equals the signal power at that scale.
We partition wavenumbers into three frequency ranges (low, mid, high) and report NEPS for each range. We compute NEPS per timeframe and per field, obtaining aggregate metrics by averaging across timeframes and fields.

\subsection{Training Objective} \label{subsec:objective}

The autoregressive rollout model predicts the temporal evolution of physical systems. Given a sequence of timesteps $\vx_{0:t-1}$, the model predicts the next frame $\hat{\vx}_t$ and is trained to minimise mean absolute error (MAE): 
\begin{align}
    \ell_\text{Rollout}
    =
    \norm{ \vx_t - \hat{\vx}_t }_1 .
\end{align}

The tokeniser is trained with an autoencoding objective, minimising mean squared error (MSE) between the reconstruction $\hat{\vx}_{0:t-1}$ and input sequence $\vx_{0:t-1}$:
\begin{align}
    \ell_\text{Tokeniser}
    =
    \norm{ \vx_{0:t-1} - \hat{\vx}_{0:t-1} }_2^2 .
\end{align}

\subsection{Architecture} \label{subsec:architecture}

Our autoregressive rollout model consists of two components: (i) a tokeniser that compresses spatiotemporal data into compact latent representations, and (ii) a transformer-based processor that operates on these tokens to predict future states. The tokeniser and processor are connected by a fully connected projection layer that matches their dimensions.

\textbf{Processor.} We use the processor architecture from Walrus \citep{mccabe2025_walrus}, which employs factorised spatial and temporal attention with axial positional encodings and a causal temporal structure. We refer readers to \citet{mccabe2025_walrus} for detailed architectural specifications.

\textbf{Tokeniser.} We use a simplified version of MAGVIT-2 \citep{yu2024_magvit2}, retaining the causal-convolutional encoder-decoder backbone but removing vector quantisation, adversarial and perceptual losses, and adaptive group normalisation. The tokeniser is trained solely with MSE reconstruction on continuous, non-quantised latents. We further extend this architecture to support runtime-adjustable spatiotemporal compression by adapting the approach of \citet{mukhopadhyay2025_flexi} to causal convolutions. This enables flexible trade-offs between compression ratio and reconstruction fidelity. Full architectural details are provided in Appendix~\ref{appendix:architecture}.

During rollout training, the processor and the projection layer are always trainable. For the tokeniser, we consider two parameter training strategies. In the \textit{fully trainable} setting, all tokeniser parameters are updated jointly with the processor. In the \textit{mostly frozen} setting, the core of the tokeniser is frozen and only the layers directly interfacing with pixel space and latent space are trainable. Specifically, this includes the encoder head, the decoder head, and the bottleneck layers, while all intermediate blocks remain frozen.

\subsection{Parameter Counts and Training Recipe} \label{subsec:training}

The tokeniser's encoder progressively increases channel dimensions from the number of physical fields to 16, 32, and 64 channels, then compresses to a bottleneck of 18 latent channels. The decoder mirrors this structure. The processor consists of six transformer blocks with embedding~dimension~1088. Table~\ref{tbl:param_counts} summarises the parameter counts for each component. Table~\ref{tbl:freezing_strategies} shows the number of trainable parameters under each parameter training strategy. In the \textit{fully trainable} setting, all tokeniser parameters are updated during rollout training. In the \textit{mostly frozen} setting, only 2\% of them remain trainable.

\begin{table}[h]
\caption[Parameter Counts]{Parameter counts for the tokeniser and downstream components.}
\centering
\label{tbl:param_counts}
\small
\centering
\begin{tabular}{c c c}
\toprule
Tokeniser & Processor \& Projection & Total \\
\midrule
5 M & 100 M & 105 M \\
\bottomrule
\end{tabular}
\end{table}

\begin{table}[h]
\caption[Parameter Training Strategies]{Parameter training strategies and corresponding trainable parameter budgets.}
\centering
\label{tbl:freezing_strategies}
\small
\begin{tabular}{l c c c}
\toprule
Strategy & Tokeniser & \# Trainable & Trainable Fraction \\
\midrule
Fully trainable (\unlockicon) & \multirow{2}{*}{5 M} & 5 M & 100 \% \\
Mostly frozen \,(\lockicon) &  & 85 k & 2 \% \\
\bottomrule
\end{tabular}
\end{table}

From each 10-frame sequence described in Subsection \ref{subsec:data}, the first 9 are used for autoencoding in tokeniser pretraining and as context for predicting the 10th frame in rollout training.
We use an effective batch size of 16 (two examples per GPU across 8 GPUs) for all experiments. We pretrain the tokeniser for 168,000 steps using the SOAP optimiser \citep{vyas2025_soap}. For rollout training, we train for 29,400 steps using the AdamW optimiser \citep{loshchilov2019_adamw} with a learning rate scheduler. When pretraining on multiple datasets, we sample each dataset with equal probability. Tokeniser pretraining uses distributed data parallel (DDP), while rollout training uses fully sharded data parallel (FSDP). All experiments are conducted on compute nodes comprising eight H100 GPUs and 64 Intel Ice Lake CPU cores. Additional hyperparameters are provided in Appendix~\ref{appendix:training}.

\section{Results} \label{sec:results}

\subsection{Training Cost}

To enable direct comparison of computational costs, Table~\ref{tbl:training_cost} reports wall-clock time for 2100 training steps, corresponding to approximately 4.6\% of the Euler dataset.

\begin{table}[h]
\caption{Wall-clock time for 2100 training steps (33,600 examples across eight GPUs).}
\centering
\label{tbl:training_cost}
\small
\begin{subtable}{0.48\textwidth}
\centering
\caption{Tokeniser}
\begin{tabular}{l c}
\toprule
Training dataset & Time (min.) \\
\midrule
Euler & $47.2 \pm 0.8$ \\
Multi & $26.04 \pm 0.27$ \\
\bottomrule
\end{tabular}
\end{subtable}
\hspace{-2em}
\begin{subtable}{0.48\textwidth}
\centering
\caption{Rollout}
\begin{tabular}{l c}
\toprule
Training dataset & Time (min.) \\
\midrule
Euler & $64.5 \pm 1.6$ \\
\bottomrule
\end{tabular}
\end{subtable}

\end{table}

\subsection{Downstream Performance}

\begin{figure}[H]
    \centering    
    \includegraphics[width=\textwidth]{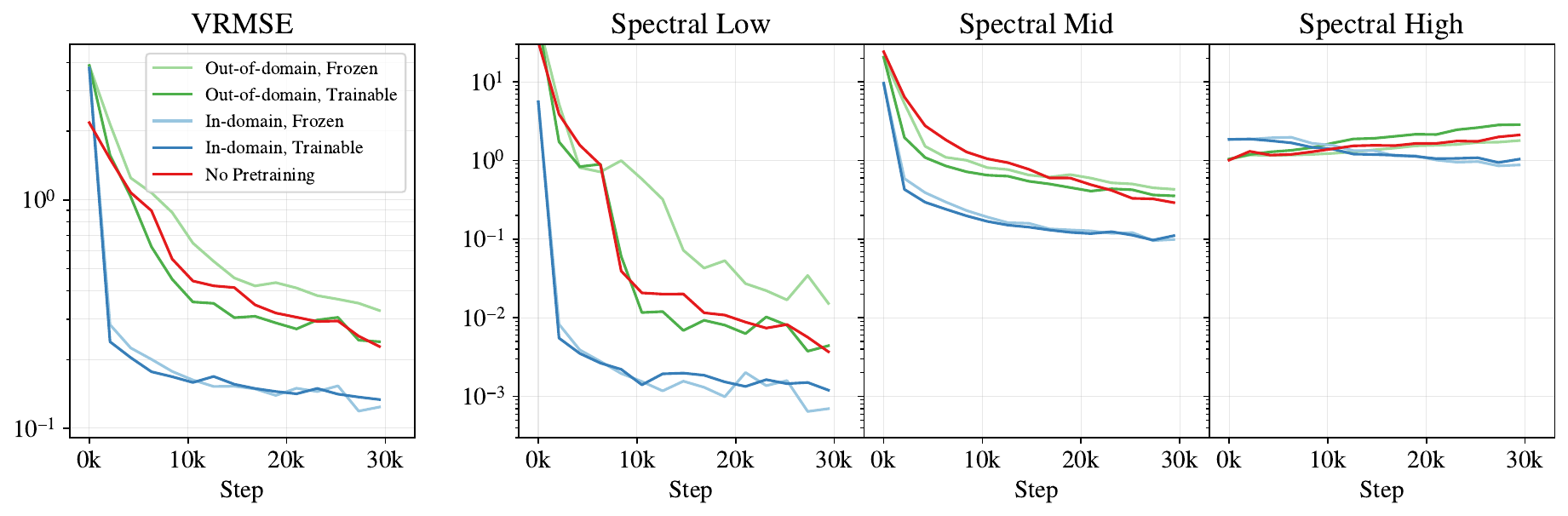}
    \caption[Validation Results]{Next-frame prediction learning curves on the validation set, shown over 29,400 training~steps for different pretraining and freezing configurations. Each panel shows a different evaluation metric: VRMSE (leftmost), and spectral error at low-, mid-, and high-frequency ranges.}
    \label{fig:validation_results}
\end{figure}

Table~\ref{tbl:downstream_performance} reports the performance of the rollout model on next-frame prediction after 10,500 training steps (168,000 examples across eight GPUs). We focus on this early training regime to assess the computational efficiency benefits of pretraining under limited budgets. All metrics are evaluated on the validation set by computing predictions for frame $t$ given context frames $0$ through $t-1$, then comparing the predicted frame $\hat{\vx}_t$ to the ground truth $\vx_t$. To reduce sensitivity to the particular choice of validation examples, we randomly sample 10 subsets of 4096 examples from the validation set, compute metrics on each subset, and report the mean across these 10 subsets. Figure~\ref{fig:validation_results} shows how these validation metrics evolve over a period of 29,400 steps for a representative run.

\begin{table}[h]
\caption{
    Next-frame prediction performance on the validation set after 10,500 training steps for~different pretraining and freezing configurations. Metrics are averaged over 10 random subsets of 4096 validation examples. The \,\unlockicon\, and \,\lockicon\, symbols indicate \textit{fully trainable} and \textit{mostly frozen} tokenisers, respectively.}
\centering
\label{tbl:downstream_performance}
\small
\begin{tabular}{l c c c c}
\toprule
& VRMSE & Spectral--Low & Spectral--Mid & Spectral--High \\
\midrule
\unlockicon\; \textcolor[HTML]{E41A1C}{\textbf{No pretraining}}
    & $0.439$
    & $0.021$
    & $1.039$
    & $1.408$ \\
\unlockicon\; \textcolor[HTML]{377EB8}{\textbf{In-domain}}
    & $0.158$
    & $0.001$
    & $0.167$
    & $1.381$ \\
\lockicon\; \textcolor[HTML]{99C6E0}{\textbf{In-domain}}   
    & $0.162$
    & $0.002$
    & $0.190$
    & $1.545$ \\
\unlockicon\; \textcolor[HTML]{4DAF4A}{\textbf{Out-of-domain}}
    & $0.355$
    & $0.012$
    & $0.642$
    & $1.659$ \\
\lockicon\; \textcolor[HTML]{A1D99B}{\textbf{Out-of-domain}}   
    & $0.643$
    & $0.544$
    & $0.801$
    & $1.229$ \\
\bottomrule
\end{tabular}
\end{table}

Tokeniser pretraining provides substantial improvements under limited computational budgets, with gains depending on domain alignment.
In-domain pretraining reduces VRMSE by 64\% compared~to training from scratch (Table~\ref{tbl:downstream_performance}: 0.158 \!vs. \!0.439 for fully trainable; 0.162 \!vs. \!0.439 when mostly frozen). Out-of-domain pretraining provides more modest gains of 19\% when the tokeniser remains trainable, though freezing it in this setting degrades performance.

Figure~\ref{fig:validation_results} reveals how pretraining affects learning dynamics across spatial scales.
In-domain pretrained models rapidly achieve the lowest VRMSE: both \textit{fully trainable} and \textit{mostly frozen} variants reach low error within 10,000 steps and maintain this advantage throughout training.
At low frequencies, in-domain models improve by more than two orders of magnitude early in training. The out-of-domain trainable and no-pretraining models eventually reach similar error levels to each other, while the out-of-domain frozen model remains at higher error.
At mid frequencies, all pretrained models initially outperform training from scratch. In-domain pretraining achieves the lowest error and maintains this advantage throughout training. 
The out-of-domain trainable model converges to the no-pretraining baseline around 20,000 steps, whereas the out-of-domain frozen model begins to underperform the baseline around 15,000 steps.
At high frequencies, all models produce poor quality (NEPS $\approx$ 1.0), but their learning dynamics differ: in-domain models show improving performance, whereas models without pretraining or with out-of-domain pretraining degrade over training.

\subsection{Freezing Strategies}

\begin{figure}[H]
    \centering    
    \includegraphics[width=\textwidth]{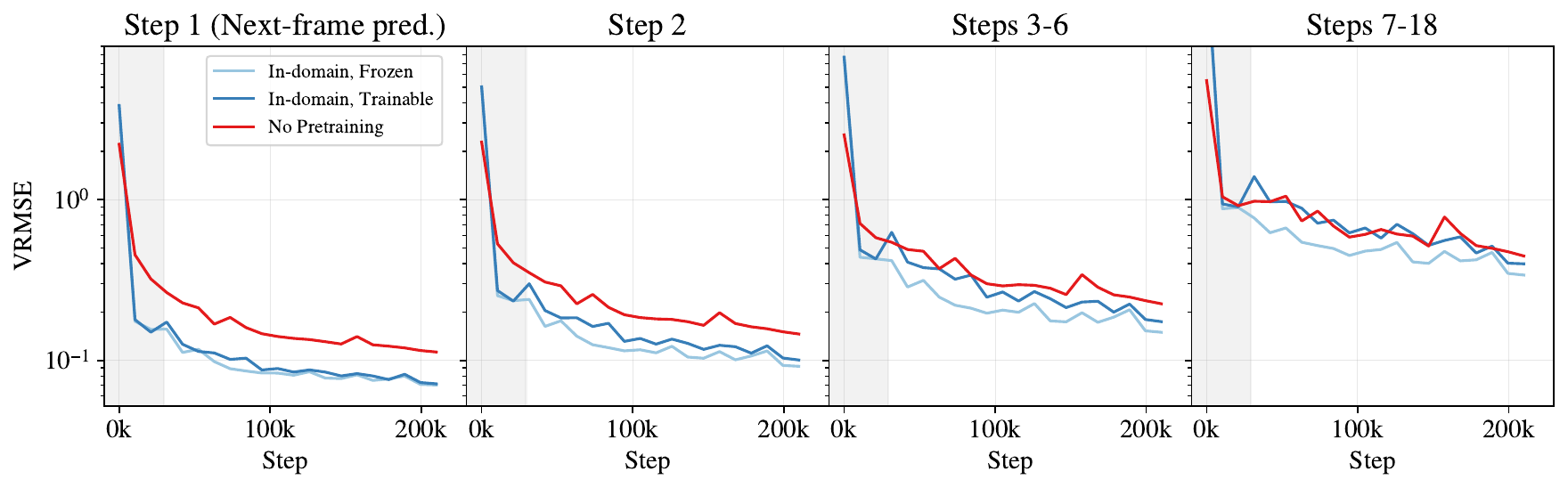}
    \caption[Rollout Validation Results]{
        Autoregressive rollout learning curves over 210,000 training steps. Each panel shows VRMSE at a different rollout horizon. The shaded region indicates the training range in Figure~\ref{fig:validation_results}.}
    \label{fig:rollout_validation_results}
\end{figure}

For the in-domain case, the \textit{fully trainable} and \textit{mostly frozen} variants achieve nearly identical performance on next-frame prediction, raising the question of whether one approach offers any~advantages. To investigate this, we examine later rollout steps, where errors can accumulate. We also extend training to 210,000 steps to assess whether differences emerge with longer training. Specifically, we evaluate autoregressive rollout quality over 18 prediction steps.\footnote{Starting from 9 frames of ground truth context ($t=0$ to $t=8$), the model autoregressively predicts the next frame ($t=9$). Each prediction is then fed back as input for the subsequent prediction. We continue this process until frame $t=26$ (18 autoregressive steps).} Figure~\ref{fig:rollout_validation_results} shows VRMSE on the validation~set across early (step 1 and step 2), medium (steps 3-6), and long (steps 7-18) rollout horizons for a representative run, with each metric averaged over 1024 validation examples.

As expected, rollout quality degrades at longer prediction horizons across all configurations, as errors accumulate over autoregressive steps. However, for in-domain pretraining, the \textit{mostly frozen} tokeniser consistently outperforms the \textit{fully trainable} variant, with the performance gap widening at longer rollout windows, most notably at steps 7-18. This suggests that freezing most of the pretrained tokeniser provides a regularisation benefit that helps prevent error accumulation during long rollouts.

\section{Discussion} \label{sec:discussion}

\subsection{Key Findings}

Our results provide the first systematic evidence on the role of tokeniser pretraining in physics foundation models.
We show that pretraining substantially accelerates autoregressive training under limited computational budgets, with benefits critically dependent on domain alignment between pretraining data and autoregressive training data. Key findings include:

\textbf{Domain alignment determines pretraining benefits.} In-domain pretraining reduces VRMSE by 64\% after 10,500 training steps compared to training from scratch, with this advantage emerging early and persisting throughout training. Out-of-domain pretraining provides more modest improvements of 19\% when the tokeniser remains trainable, though freezing it in this setting degrades performance below the no-pretraining baseline.

\textbf{Scale-dependent learning dynamics.} Pretraining affects learning differently across spatial scales (Figure~\ref{fig:validation_results}). At low frequencies, in-domain models achieve excellent final performance (NEPS~$\sim$~0.001) compared to models without pretraining (NEPS~$\sim$~0.02). At mid frequencies, in-domain models maintain consistently lower error throughout training (NEPS~$\sim$0.1--0.3 vs. $\sim$0.3--1.0 without pretraining). At high frequencies, all models show poor quality (NEPS $\approx$ 1.0), but in-domain pretraining enables continuous improvement whereas other models show progressive degradation.

\textbf{Frozen tokenisers can outperform trainable ones for long rollouts.} In-domain pretrained models perform similarly on next-frame prediction whether the tokeniser is fully trainable or mostly frozen. However, the frozen variant consistently outperforms on longer autoregressive rollouts, with its advantage widening at longer horizons. This suggests freezing the pretrained tokeniser provides regularisation that prevents error accumulation. Practically, a mostly frozen tokeniser reduces trainable parameters by 98\% (from 5M to 85k) while improving rollout quality.

\subsection{Limitations}

Several limitations warrant discussion.
First, out-of-domain pretraining provides modest benefits, and freezing the tokeniser in this setting degrades performance below the no-pretraining baseline. One explanation is limited field overlap between pretraining and downstream datasets: only one of the four physical quantities in the Euler dataset appears in the multi-dataset mixture (Table~\ref{tbl:data}), while the others are encountered only during rollout training. Future work could investigate whether greater field overlap improves out-of-domain transfer, or develop methods to help pretrained tokenisers adapt to previously unseen physical fields.

Second, while our results demonstrate clear efficiency gains from pretraining, absolute error values remain higher than state-of-the-art physics emulators. Additionally, all models show limited prediction quality at high spatial frequencies. These limitations likely stem from the constrained capacity of our compact 5M-parameter tokeniser. Scaling studies with larger models, as well as architectural modifications, are needed to reach competitive performance levels and preserve fine-scale structure.

Finally, although we introduce flexible spatiotemporal compression and its theoretical motivation, its practical benefits have yet to be thoroughly investigated. Systematic experiments examining how flexible compression affects both autoencoding quality and downstream performance would strengthen our understanding of this capability and its advantages over fixed compression.

\subsection{Future Directions}

\textbf{Downstream task diversity.} While this work focuses on autoregressive rollout prediction, pretrained tokenisers may prove valuable across a broader range of downstream settings.
Applying the same pretrained tokeniser to diverse tasks and training objectives -- such as masked token modelling, physical parameter inference, or inverse problems -- would help reveal which settings benefit most from autoencoding pretraining.

\textbf{Tokeniser architecture \& training improvements.} Investigating tokeniser design could yield more general insights and improve both pretraining and downstream accuracy. Promising directions include: evaluating alternative architectures to identify inductive biases better suited to physics data; scaling studies to determine whether benefits compound at larger capacity and with more diverse pretraining data; comparing training objectives (reconstruction, perceptual, adversarial losses) and quantisation strategies; and determining optimal pretraining budgets (measured in training steps, dataset size, or target accuracy) to maximise downstream performance while minimising computational cost.

\textbf{Data composition \& domain alignment.} Our results show that domain alignment between pretraining and downstream datasets significantly impacts performance gains, raising an important question about data composition: given a fixed pretraining budget, does training on diverse multi-physics mixtures produce more versatile and generalisable tokenisers than domain-specific pretraining? Furthermore, systematically varying domain similarity, from small differences (boundary conditions, physical parameters) to large ones (governing equations), would quantify how domain proximity impacts transfer.
Additionally, varying the size of the downstream dataset would reveal how pretraining gains scale with data availability and whether tokeniser pretraining is especially valuable in data-limited regimes.

\textbf{Physics-aware evaluation metrics.} While our spatial and spectral metrics assess reconstruction fidelity, they do not capture whether models respect underlying physical principles. Developing physics-informed metrics that assess conservation laws, symmetries, and domain-specific constraints would provide deeper insight into model quality. Extending spectral analysis to the temporal domain would further reveal how models capture dynamics across timescales, identifying whether they accurately represent both fast transients and slow evolution or systematically bias toward certain temporal frequencies.

Together, these research directions would yield principled guidelines for tokeniser pretraining in physics foundation models, clarifying when pretraining justifies its computational cost and how data composition and domain alignment shape its benefits.
Additionally, if future investigations demonstrate consistent out-of-domain benefits, they would justify developing and releasing high-quality pretrained tokenisers as community resources across scientific and engineering domains.

\section{Conclusion} \label{sec:conclusion}

This work provides the first systematic investigation of tokeniser pretraining for physics foundation~models. We demonstrate that pretraining offers a practical path to improved training efficiency, with benefits that depend critically on domain alignment between pretraining and downstream datasets. We also find that freezing most of an in-domain pretrained tokeniser can improve long-horizon \linebreak[4] rollouts, suggesting a simple regularisation strategy for stable autoregressive prediction. These findings position tokeniser pretraining as a simple yet effective mechanism for improving the training of physics foundation models.

Looking forward, the strong dependence on domain alignment raises important questions about optimal pretraining strategies. As physics foundation models continue to scale and tackle increasingly complex problems, understanding when and how to leverage pretrained tokenisers will be essential for their effective deployment across diverse computational settings. The insights from this work contribute to a broader understanding of how to build efficient, scalable foundation models that can accelerate scientific progress across domains.


\subsubsection*{Acknowledgments}

The authors thank Géraud Krawezik and the Scientific Computing Core at the Flatiron Institute, a division of the Simons Foundation, for providing computational facilities and technical support. This work made use of compute resources provided by the Flatiron Institute.
Work conducted at the University of Cambridge was carried out within the Institute of Astronomy, the Kavli Institute for Cosmology, the Department of Computer Science and Technology, and the Department of Applied Mathematics and Theoretical Physics.
Polymathic\,AI gratefully acknowledges funding from the Simons Foundation and Schmidt Sciences, LLC.

Hadi Sotoudeh acknowledges support from the Centre for Doctoral Training in Data Intensive~Science and is grateful to Girton College, where he is an Emily Davies Scholar.
Payel Mukhopadhyay thanks the Infosys–Cambridge AI Centre for support.
Shirley Ho acknowledges support from the~Schmidt~Sciences AI2050 Senior Fellowship.
Miles Cranmer acknowledges support from the~Schmidt~Sciences AI2050 Early Career Fellowship and the Isaac Newton Trust.
We thank Lucas~Meyer for research engineering support, Rudy Morel, François Lanusse, Helen Qu, and Tom~Hehir for valuable discussions and feedback, and François Rozet for assistance with setting up the SOAP optimiser. We also thank the Polymathic\,AI and Astro\,Automata teams for helpful discussions and feedback throughout this project, and the reviewers for their valuable comments.


\newpage
\bibliography{iclr2026_conference}
\bibliographystyle{iclr2026_conference}

\appendix

\clearpage
\section{Dataset Properties} \label{appendix:data}

\vspace{0.1\baselineskip}
\paragraph{Euler multiquadrants.}
Evolution of different gases starting with piecewise constant initial data in quadrants. The evolution can give rise to shocks, rarefaction waves, contact discontinuities, interaction with each other and domain walls. The evolved fields are density $\rho$, energy $e$, pressure~$p$, and momentum components $(p_x, p_y)$, and the system follows Euler equations for a compressible gas. The data are sampled on a uniform 512$\times$512 Cartesian grid over 100 timesteps. The dataset comprises 20 parameter configurations, varying in boundary condition type (open or periodic) and gas constant~$\gamma$, with 500 trajectories per configuration.

\paragraph{Rayleigh-Bénard convection.}
Simulations of classical Rayleigh-Bénard convection, where a horizontal fluid layer heated from below develops convective rolls (Bénard cells) due to thermal buoyancy forces. The evolved fields include buoyancy $b$, pressure $p$, and velocity components $(v_x, v_y)$, governed by the Boussinesq approximation of the Navier-Stokes equations. The data are sampled on a 512$\times$128 Cartesian grid over 200 timesteps, across 35 parameter sets and 50 trajectories per parameter set.

\paragraph{Shear flow.}
Continuous deformation of adjacent fluid layers sliding past each other with different velocities. The evolved fields are tracer $s$, pressure $p$, and velocity components $(v_x, v_y)$, and the system is governed by incompressible Navier-Stokes equations. The data are sampled on a uniform 256$\times$512 Cartesian grid over 200 timesteps, across 28 flow configurations and 40 trajectories per configuration.

\paragraph{Active matter.}
Simulation of a continuum model for biological active matter, comprising rod-like particles immersed in a Stokes fluid. The evolved fields include a tracer-like concentration field~$c$, fluid velocity $(v_x, v_y)$, an orientation tensor (describing alignment of the active agents), and a strain-rate tensor. The data are sampled on a uniform 256$\times$256 Cartesian grid over 81 timesteps, across 45 parameter sets and 5 trajectories per parameter set.

\clearpage
\section{Tokeniser Architecture and Flexible Compression} \label{appendix:architecture}

\begin{figure}[H]
    \centering    
    \includegraphics[width=0.95\textwidth]{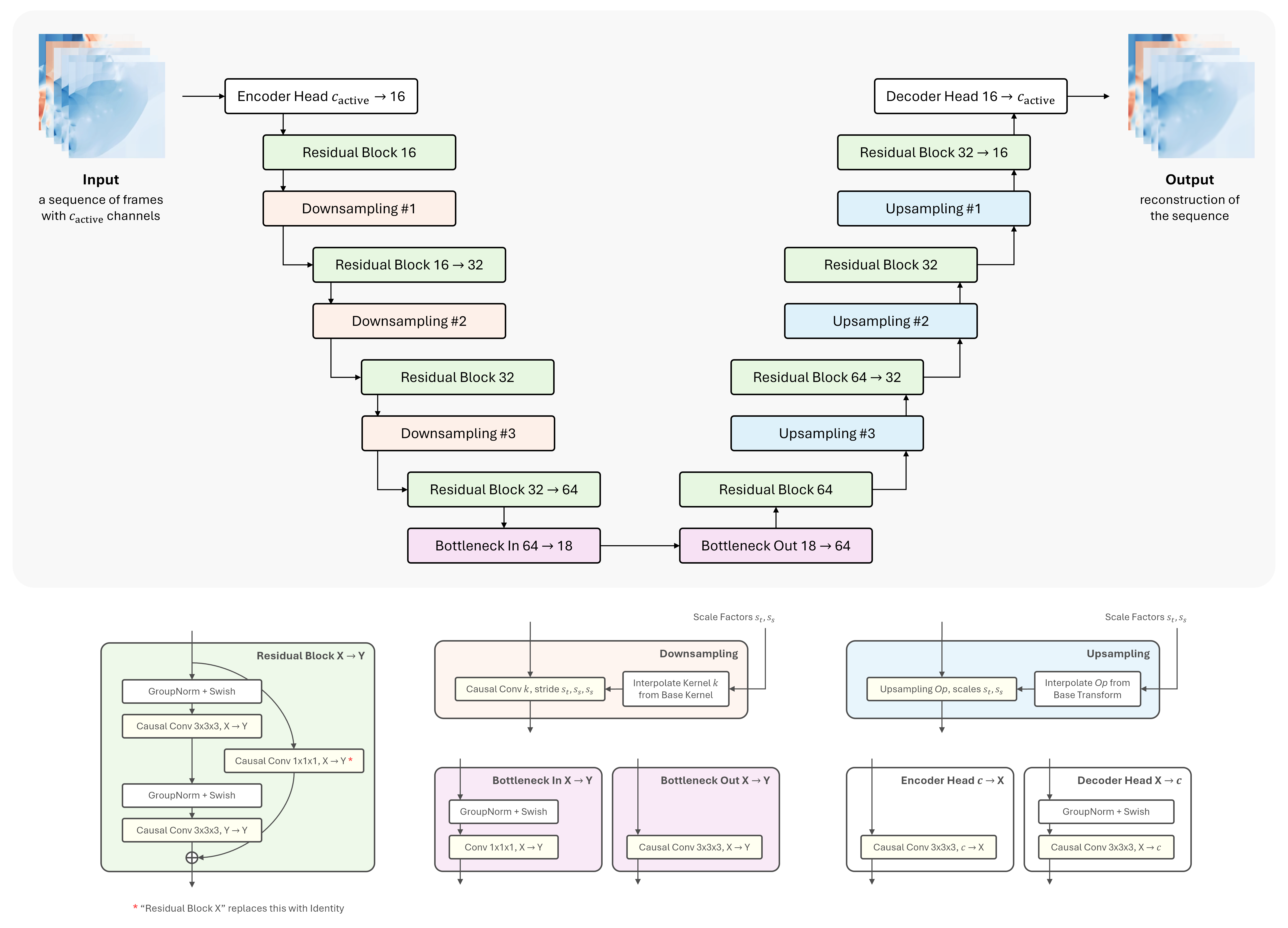}
    \caption[Architecture]{MAGVIT-Simple architecture.}
    \label{fig:architecture}
\end{figure}

MAGVIT-Simple autoencodes temporal sequences of 2D or 3D simulations $\vx \in \sR^{T \times S_1 \times \dots \times S_k}$, where $k \in \{2, 3\}$. The temporal size $T$ is set to ``1 + a power of two'', and the spatial sizes $S_i$ are powers of two. Figure~\ref{fig:architecture} illustrates the tokeniser architecture.

We employ causal convolutions, as introduced in MAGVIT-2, which enable joint image-video tokenisation and have been shown to outperform non-causal convolutions and attention-based blocks~\citep{yu2024_magvit2}. For 2D simulations, a 3D causal convolutional layer with kernel size $(k_t, k_s, k_s)$ is applied to the input.\footnote{Operating on 3D simulations would require 4D causal convolutions.} Spatial padding is applied symmetrically to both ends of each spatial dimension and is configured to preserve the input’s spatial size. Temporal padding is applied only at the beginning (i.e., $k_t - 1$ leading frames), with no padding at the end. This preserves the temporal size of the input and ensures that each output timeframe depends only on the current and preceding frames. Padding values are set to zero.

The encoder and decoder heads use adaptive causal 3D convolutions that support a variable number of input physical fields. A single convolutional kernel is learned for the maximum number of fields,~$c_\text{total}$. At runtime, given an input with $c_\text{active}$ fields, rows and columns corresponding to active fields are selected from the weight matrix, and the output is rescaled by $\sqrt{c_{\text{total}} / c_{\text{active}}}$ to compensate for reduced fan-in. This follows the field embedding approach from MPP~\citep{mccabe2024_mpp}.

To enable flexible downsampling, we adapt the convolutional kernel modulation of \citet{mukhopadhyay2025_flexi} to causal convolutions. For downsampling, we follow the idea of convolutional kernel modulation introduced in \citet{mukhopadhyay2025_flexi} and extend it to causal convolutions. Let the set of scaling factors in a downsampling layer be denoted by $\{\, (s_t, s_s)_i \,\}$, where each pair $(s_t, s_s)$ consists of temporal and spatial strides, both chosen as powers of two. We use strided causal convolutions with adaptive kernel sizes to perform downsampling.

Given a scaling pair $(s_t, s_s)$, we define the spatial and temporal kernel sizes as:
\begin{equation}
    k_s = s_s
    \qquad \qquad \qquad
    k_t =
    \begin{cases}
        1+s_t & \text{if } s_t > 1, \\
        s_t & \text{if } s_t = 1.
    \end{cases}
    \;.
\end{equation}
The stride is set to the corresponding scaling factor. This results in non-overlapping convolutions in space and a one-frame overlap in time. The kernel weights are obtained by interpolating from a base kernel, corresponding to the largest scaling factors $(s_{t,\text{max}}, s_{s,\text{max}})$ and thus having the largest size.

For flexible upsampling, we explore two strategies: causal transposed convolutions and the depth-to-space (pixel shuffle) operation. The former follows the same principle as flexible strided convolutions, with an additional convolution at the end to adjust the output dimensions. In the latter approach, the number of channels is first expanded by the product of the scaling factors, $s_t \times s_s^k$, via a causal convolution. The resulting tensor is then rearranged, redistributing channel values into the spatiotemporal dimensions.

To enable flexible upsampling with depth-to-space, we define the base convolutional kernel that upsamples by the largest scaling factor. For any smaller scaling factor that divides the base, we first perform base upsampling and then subsample the output to achieve the desired lower dimensionality. In some cases, this behaviour can be reproduced more efficiently by subsampling the base convolutional kernel directly, thereby avoiding redundant computation. One such case arises when the subsampling strategy selects every
\begin{align*}
    \eta =
    \left(
        \frac {s_{t, \text{base}} } {s_t}
        \; , \;\;
        \frac {s_{s, \text{base}} } {s_s}
    \right)
\end{align*}
pixel in the output; a strategy we adopt in our flexible depth-to-space layers. Finally, because this operation introduces additional timeframes, we follow MAGVIT-2 \citep{yu2024_magvit2} and discard the first $s_t - 1$ frames from the output.

\clearpage
\section{Detailed Training Recipe} \label{appendix:training}

This appendix provides the complete set of hyperparameters for both training stages. High-level training details are described in Subsection~\ref{subsec:training}, and the tokeniser architecture is described in Appendix~\ref{appendix:architecture}. Table~\ref{tbl:hyperparams} summarises the training hyperparameters.


\begin{table}[h]
\caption{Training hyperparameters for tokeniser pretraining and autoregressive rollout training.}
\centering
\label{tbl:hyperparams}
\small
\begin{tabular}{l c c}
\toprule
Hyperparameter & Tokeniser Pretraining & Rollout Training \\
\midrule
\multicolumn{3}{l}{\textit{Optimiser}} \\
\quad Optimiser & SOAP & AdamW \\
\quad [Peak] learning rate & $5 \times 10^{-4}$ & $5 \times 10^{-5}$ \\
\quad Betas & $(0.95,\, 0.95)$ & $(0.9,\, 0.999)$ \\
\quad Weight decay & $0.01$ & $10^{-4}$ \\
\quad $\epsilon$ & $10^{-8}$ & $10^{-10}$ \\
\quad LR scheduler & --- & Inv.\ sqrt (see below) \\
\midrule
\multicolumn{3}{l}{\textit{Training}} \\
\quad Max steps & 168,000 & 29,400 \\
\quad Gradient clipping & 0.5 & 5.0 \\
\quad Distribution strategy & DDP & FSDP \\
\bottomrule
\end{tabular}
\end{table}


\paragraph{Learning rate schedule.}
For tokeniser pretraining, no learning rate scheduler is used; the learning rate remains constant at $5 \times 10^{-4}$ throughout training.

For rollout training, we use an inverse-square-root schedule with linear warmup and square-root cooldown, stepping once per epoch. Let $e$ denote the current epoch (0-indexed), with total epochs~$E$, warmup duration $W$, cooldown duration $C$, warmup starting factor $\alpha = 0.1$, and decay offset $\kappa = 128$. The schedule has three phases:

\noindent\textit{Warmup} ($e \leq W$):\quad The learning rate increases linearly:
\begin{equation}
\text{lr}(e) = \text{lr}_\text{peak} \left(\alpha + (1 - \alpha) \frac{e}{W}\right).
\end{equation}

\vspace{0.5\baselineskip}
\noindent\textit{Decay} ($W < e \leq E - C$):\quad The learning rate follows an inverse-square-root decay:
\begin{equation}
\text{lr}(e) = \frac{\text{lr}_\text{peak}}{1 + \sqrt{s + \kappa} - \sqrt{\kappa}}\,,
\end{equation}
where $s := e - W$ counts epochs into decay ($s = 1, 2, \ldots$).

\vspace{0.5\baselineskip}
\noindent\textit{Cooldown} ($e > E - C$):\quad The learning rate ramps down with a square-root profile:
\begin{equation}
\text{lr}(e) = \text{lr}_{\text{end}} \left(1 - \sqrt{\frac{t - 1}{C}}\right),
\end{equation}
where $t := e - (E - C)$ counts epochs into cooldown ($t = 1, 2, \ldots$) and
\begin{equation}
\text{lr}_{\text{end}} := \frac{\text{lr}_\text{peak}}{1 + \sqrt{E - C + \kappa} - \sqrt{\kappa}}\,.
\end{equation}


\paragraph{Data loading and recycling.}
Each GPU processes a disjoint shard of the training data. For tokeniser pretraining, each shard contains 21{,}000 unique batches (of size~2). Once all unique batches have been consumed, the data loader reseeds its random number generator and begins iterating over the dataset again, allowing resampling in subsequent passes. For rollout training, the same mechanism applies but each shard contains 2100 unique batches.

\paragraph{Convolution details.}
Encoder and decoder heads use causal 3D convolutions with kernel size $(k_t, k_s, k_s) = (3, 3, 3)$ and stride 1. Temporal padding is applied only at the past (leading $k_t - 1 = 2$ frames) so that each output timestep depends only on current and previous inputs; spatial padding is symmetric (1 pixel per side). All residual blocks use causal 3D convolutions with kernel size 3 and stride 1. The encoder bottleneck uses a $1 \times 1 \times 1$ convolution to project to the latent dimension; the decoder bottleneck uses kernel size 3.

In downsampling layers, stride equals the chosen compression $(s_t, s_s, s_s)$ (see Table~\ref{tbl:compressions}). The convolution kernel size is tied to the stride: temporal size is $k_t = 1 + s_t$ when $s_t > 1$ and $k_t = s_t$ when $s_t = 1$; spatial size is $k_s = s_s$. Thus for the flexible layers the temporal kernel size is 1 or 3 and the spatial kernel size is 2 or 4. Spatial padding in these layers is ``valid'' (no padding).

We use depth-to-space upsampling. The corresponding layers employ causal $3 \times 3 \times 3$ convolutions. These convolutions follow the same padding scheme as elsewhere: causal in time (two leading frames, no trailing) and symmetric in space (one pixel on each side) so that spatial size is preserved before the depth-to-space rearrangement.


\paragraph{Compression factors.}
Table~\ref{tbl:compressions} summarises the temporal and spatial compression factors at each downsampling depth of the tokeniser (see Appendix~\ref{appendix:architecture} for the flexible compression mechanism). During training, the tokeniser uniformly samples one factor from each set at every forward pass; during validation, the scales are fixed to $(1, 2),\, (2, 2),\, (2, 2)$ for depths \#1--\#3 respectively.

\begin{table}[h]
\caption{Temporal and spatial compression factors at each tokeniser depth. Values in braces indicate the set from which a factor is uniformly sampled during training.}
\centering
\label{tbl:compressions}
\small
\begin{tabular}{l c c}
\toprule
Depth & Temporal Compression & Spatial Compression \\
\midrule
\#1 & 1 & 2  \\
\#2 & \{1, 2\} & \{2, 4\} \\
\#3 & \{1, 2\} & \{2, 4\}  \\
\bottomrule
\end{tabular}
\end{table}


\paragraph{Processor configuration.}
As noted in Subsection~\ref{subsec:training}, the processor has an embedding dimension of 1088 and the tokeniser bottleneck dimension is 18. A linear projection maps between the 18-dimensional tokeniser output and 1088-dimensional processor input, with a symmetric projection mapping back after the final block. Each of the six processor blocks consists of full spatial attention followed by axial temporal attention, both using 16 attention heads. Spatial attention employs a SwiGLU MLP with $4\times$ expansion ratio (hidden dimension 4352) and Rotary Position Embeddings (RoPE) with axial frequencies. Temporal attention uses causal masking and learned relative position biases. Block normalisation is RMSGroupNorm and stochastic depth (drop path) is applied with a linearly increasing rate from 0 to 0.05 across blocks.


\paragraph{Input normalisation.}
In both training stages, each input sample is normalised before being fed to the tokeniser. We compute the root mean square (RMS) of each physical field over all timesteps and spatial locations, and divide by this value (clamped below at $\epsilon = 10^{-7}$ for numerical stability). The inverse transform is applied to the output. This samplewise normalisation operates independently per sample and per field.

\end{document}